\documentclass[conference]{IEEEtran}
\IEEEoverridecommandlockouts

\usepackage[noadjust]{cite}
\usepackage{amsmath,amssymb,amsfonts}
\usepackage{algorithmic}
\usepackage{graphicx}
\usepackage{textcomp}
\usepackage{xcolor}
\usepackage{lipsum}
\usepackage{tikz}
\usepackage{pgfplots}
\usepackage{booktabs,siunitx}
\usepackage[verbose,margin=3cm]{geometry}
\usepackage[acronym]{glossaries}

\usepackage[T1]{fontenc}
\usepackage{newtxtext,newtxmath}
\usepackage[utf8]{inputenc}
\usepackage[english]{babel}
\usepackage[breaklinks]{hyperref}
\usepackage{multirow}
\usepackage{cleveref}
\usepackage{ragged2e}

\newcolumntype{T}[1]{S[table-format=#1]}
\def\BibTeX{{\rm B\kern-.05em{\sc i\kern-.025em b}\kern-.08em
    T\kern-.1667em\lower.7ex\hbox{E}\kern-.125emX}}

\newacronym{ann}{ANN}{artificial neural network}
\newacronym{snn}{SNN}{spiking neural network}
\newacronym{dvs}{DVS}{dynamic vision sensors}
\newacronym{roi}{ROI}{region of interest}
\newacronym{ssm}{SSM}{state-space model}
\newacronym{gru}{GRU}{gated recurrent unit}
\newacronym{cnn}{CNN}{convolutional neural network}
\newacronym{lif}{LIF}{leaky integrate-and-fire}
\newcommand\x{2.0cm}

\pgfplotsset{compat=1.18}
    
\begin{document}

\title{Neuromorphic Eye Tracking for Low-Latency Pupil Detection
\thanks{NimbleAI, Edgy Organism, NASCENT2.0}
}

\author{
\IEEEauthorblockN{Paul Hueber}
\IEEEauthorblockA{
\textit{University of Manchester, UK}\\
paul.hueber@manchester.ac.uk}
\and
\IEEEauthorblockN{Luca Peres}
\IEEEauthorblockA{
\textit{University of Manchester, UK}\\
luca.peres-2@manchester.ac.uk}
\and
\IEEEauthorblockN{Florian Pitters}
\IEEEauthorblockA{\textit{Viewpointsystem GmbH} \\
f.pitters@viewpointsystem.com}
\and
\hspace{\x}\IEEEauthorblockN{Alejandro Gloriani}
\hspace{\x}\IEEEauthorblockA{$\qquad\qquad\qquad\quad$\textit{Viewpointsystem GmbH} \\
\hspace{\x}a.gloriani@viewpointsystem.com}
\and
\IEEEauthorblockN{Oliver Rhodes}
\IEEEauthorblockA{
\textit{University of Manchester, UK}\\
oliver.rhodes@manchester.ac.uk}
}

\maketitle

\begin{abstract}
Eye tracking for wearable systems demands low latency and milliwatt-level power, but conventional frame-based pipelines struggle with motion blur, high compute cost, and limited temporal resolution. Such capabilities are vital for enabling seamless and responsive interaction in emerging technologies like augmented reality (AR) and virtual reality (VR), where understanding user gaze is key to immersion and interface design. Neuromorphic sensors and spiking neural networks (SNNs) offer a promising alternative, yet existing SNN approaches are either too specialized or fall short of the performance of modern ANN architectures.
This paper presents a neuromorphic version of top-performing event-based eye-tracking models, replacing their recurrent and attention modules with lightweight LIF layers and exploiting depth-wise separable convolutions to reduce model complexity. Our models obtain 3.7-4.1px mean error, approaching the accuracy of the application-specific neuromorphic system, Retina (3.24px), while reducing model size by $\mathbf{20\times}$ and theoretical compute by $\mathbf{850\times}$, compared to the closest ANN variant of the proposed model. 
These efficient variants are projected to operate at an estimated 3.9-4.9~mW with 3~ms latency at 1~kHz. The present results indicate that high-performing event-based eye-tracking architectures can be redesigned as SNNs with substantial efficiency gains, while retaining accuracy suitable for real-time wearable deployment. 
\end{abstract}


\section{Introduction}

Humans are unique among primates in having evolved eyes with a bright, uniformly white sclera, a feature that enhances the visibility of gaze direction and is fundamental to non-verbal communication \cite{kano2022experimental, wacewicz2022adaptive}. The ability to discern gaze reveals focus and intent, a capability that modern eye-tracking technology seeks to automate and quantify. Eye tracking, the process of measuring eye movements and positions, is thus a crucial method for understanding an individual's state and enabling interaction. It comprises critical metrics such as gaze direction, pupil diameter, and saccadic dynamics, which are essential for applications ranging from medical and psychological diagnostics \cite{leigh2015neurology, harezlak2018application} to human-computer interfaces in augmented and virtual reality (AR/VR) \cite{Moreno-Arjonilla2024-ld, angelopoulos2020event}, robotics \cite{Fischer-Janzen2024-xp, 9879432}, and driver monitoring systems \cite{Fan2021-xe}.

A core aspect of eye-tracking applications is pupil detection, the accurate and precise localization and tracking of the pupil within the visual field. This process enables the derivation of critical metrics such as gaze direction, blink rate, fixation duration, pupil diameter, and saccadic movement, forming the foundation for higher-level analyses. However, this task is exceptionally challenging due to the physiological characteristics of the human eye. As the fastest-moving organ in the human body, it can execute saccades of $15^\circ$ amplitude with peak velocity of $500^\circ\mathrm{s}^{-1}$ \cite{bahill1975glissades} with accelerations surpassing \(24,000^\circ\mathrm{s}^{-2}\) \cite{abrams1989speed}.

\begin{figure*}[htb!]
    \centering
    \includegraphics[width=\textwidth]{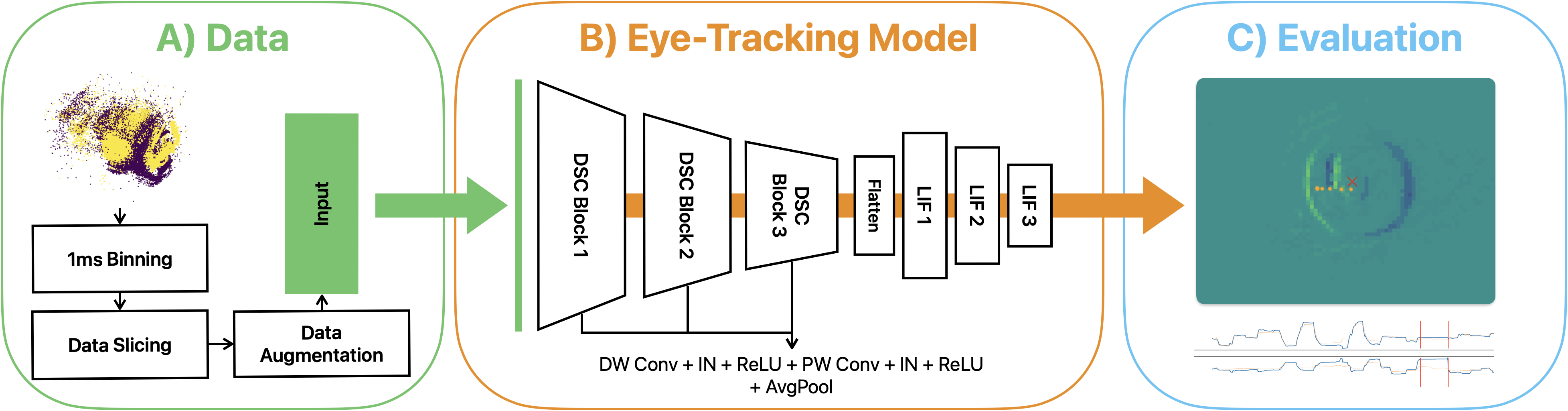}
    \caption{(A) Raw event streams are processed into 1 ms bins and sliced into 450 ms time windows for training and a continuous stream of 1 ms bins for evaluation and testing. The binned window is augmented using spatial and temporal transformations and CutOut. (B) Models are trained on the 450 ms sequences with a sliding window. (C) The best model from validation is evaluated in a continuous, online setting on the test set to simulate real-world deployment and report final performance.}
    \label{fig:pipeline}
\end{figure*}

Conventional eye-tracking systems face fundamental limitations stemming from both their sensing and processing components. Frame-based cameras inherently struggle with motion blur during rapid saccadic movements \cite{rayner2009eye}, and face a critical trade-off between temporal resolution and energy efficiency. While high-speed cameras operating at kHz rates can mitigate latency issues, they do so at the cost of increasing power consumption \cite{mele2012gaze}, rendering them impractical for energy-constrained wearable systems \cite{chen2025event}. Complementing these sensor limitations, conventional \glspl{ann} introduce additional bottlenecks, as their computational intensity and sequential frame-processing paradigm struggle to maintain low-latency inference at high throughput. This combination of hardware and algorithmic constraints typically limits practical systems to lower operational frequencies. For context, even demanding appreciations like foveated rendering require tracking updates at display refresh rates (90-120 Hz) with minimal latency to avoid user discomfort, a challenging target for power-constrained systems using conventional approaches. 

Event-based cameras, also known as \gls{dvs}, represent a paradigm shift in visual sensing with the potential to addressing limitations \cite{lichtsteiner2008128, iddrisu2024event}. Unlike conventional cameras that capture full frames at fixed intervals, \gls{dvs} can detect per-pixel brightness changes asynchronously, generating sparse event streams with microsecond temporal resolution, high dynamic range, and minimal motion blur \cite{8334288, gallego2020event}.
This makes them particularly well-suited for pupil tracking, as they focus computational resources on dynamic eye movements while ignoring static regions \cite{iddrisu2024event, angelopoulos2020event}. 

Similarly, \glspl{snn} provide a biologically-inspired processing framework that aligns well with the characteristics of event-based sensing. \glspl{snn} leverage temporal dynamics through spike-based, event-driven computation, allowing for more efficient processing of spatio-temporal event patterns compared to traditional \glspl{ann} \cite{eshraghian2023training}. When implemented on specialized neuromorphic hardware, \glspl{snn} can exploit their inherent sparsity, leading to significant gains in both latency and energy efficiency, directly addressing the challenges of conventional eye-tracking systems \cite{davies2021advancing}.

It is surprising that neuromorphic approaches have seen little use in wearable eye tracking, given their promise of inherent recurrence, improved latency and milliwatt power consumption. The \textit{Retina} demonstrated an \gls{snn} implementation with a pupil tracking error of 3.24px, power consumption between 2.89-4.8 mW, and latency of 5.57-8.01 ms \cite{bonazzi2024retina}. This falls short of the accuracy performance of \gls{ann} decoders and its continuous resetting of neuron states may disrupt continous tracking on neuromorphic hardware.
Jiang et al. \cite{jiang2024eye} presented an eye-tracking solution combining \glspl{snn} with event cameras that allows self-adjustment of time resolution with a maximum achievable resolution of 0.081 ms, but had a trade-off between spatial and temporal resolution.

This work introduces neuromorphic versions of top-performing AIS 2024 Challenge models (MambaPupil, CETM and BRAT) \cite{wang2024event}, replacing their recurrent and attention blocks with efficient LIF layers, and using depth-wise separable convolutions to require $\sim$6$\times$ fewer parameters and $\sim$82$\times$ fewer operations. Evaluated in continuous 1 kHz operations with 1 ms windows (\Cref{fig:pipeline}), the presented models achieve competitive accuracy on the 3ET+ dataset. This higher sampling rate was chosen to reflect the needs of vision scientists for high temporal resolution for precise sampling of saccades and micro-saccades \cite{raju2025evaluating, angele2025low}. Hardware projections demonstrate milliwatt-level power consumption and <8 ms latency, establishing a viable pathway for deploying complex eye-tracking on resource-constrained wearable devices. 
%
Our specific contributions include:
\begin{itemize}
\item We introduce a systematic method for converting recurrent and attention-based temporal modules from event-based ANNs into lightweight LIF layers, preserving temporal modelling capabilities at a fraction of the computational cost.
\item We demonstrate that these neuromorphic models, evaluated in a continuous 1 kHz setting, achieve comparable performance with specialized neuromorphic systems (i.e., Retina) while maintaining the flexibility of general-purpose ANN backbones.
\item We achieve a 30-$1000\times$ reduction in theoretical compute and a 22-$45\times$ model size reduction, projecting sub-5 mW power consumption, thereby establishing a new operating point for high-speed, always-on pupil-tracking.
\end{itemize}

\section{Related Works}

%

%

\subsection{Frame-based eye tracking}

Traditional eye-tracking systems rely on frame-based cameras and are primarily built upon two methodological families, model-based and appearance-based approaches, which, while distinct, are frequently combined to improve robustness and accuracy.

\subsubsection{Model-based methods}

%

Model-based approaches extract geometric features (pupil center, corneal glints) and fit them to physiological eye models for gaze estimation \cite{mestre2018robust, 6826500}. 
While these methods can achieve high accuracy, they are notoriously sensitive to deployment conditions. Their performance is heavily dependent on consistent ambient lighting \cite{gibaldi2017evaluation}, high image resolution \cite{ding2025facet}, and precise user-specific calibration \cite{geisler2018real}. Furthermore, they struggle with occlusions from eyelids and eyelashes, as well as deformations in eye shape \cite{mestre2018robust}. 

\subsubsection{Appearance-based methods}
%

Appearance-based methods typically use \glspl{cnn} for end-to-end regression from eye images to gaze points, offering robustness to shape and illumination variations. However, they demand large datasets, incur high computational costs, and are ultimately bounded by the frame rates and resolution of the input camera \cite{zhang2017mpiigaze, mazzeo2021deep, lee2020deep,deane2023deep, kim2019nvgaze}.


Both paradigms are constrained by physical camera limitations: standard high-speed cameras peak at 10-300 Hz, which is insufficient to accurately capture the fastest saccadic eye movements without motion blur. At the same time, kHz-rate cameras consume watts of power \cite{ding2025facet}, exceeding mobile power budgets and motivating event-based alternatives.

\subsection{Event-based eye tracking}



The asynchronous nature of event cameras enables microsecond temporal resolution, minimal motion blur, high dynamic range (>120 dB), minimal latency and drastically lower power consumption \cite{gallego2020event}, making them ideal for capturing rapid eye movements in power-constrained devices \cite{chen2025event}.

\subsubsection{Hybrid statistical methods}
Hybrid approaches combine event streams with occasional frames, using frames for model initialization and events for high-frequency updates. Angelopoulos et al.\cite{angelopoulos2020event} achieved >10 kHz 
with a parametric pupil model, but suffered from noise sensitivity and required recalibration. EV-EYE \cite{zhao2023ev} achieved a tracking frequency of 38.4 kHz via pupil centroid tracking, but assumed fixed pupil shape between frames. Feng et al.  \cite{feng2022real} used software-emulated events and \gls{roi} prediction for efficiency, but operated at only 30 Hz.

%

%

These methods demonstrate high-frequency tracking potential but share limitations: complex data synchronization, computational overhead from fusion of event and frames
, and persistent power consumption from frame-based components. These inherent limitations motivate exploring fully event-based approaches that eliminate the frame bottleneck entirely.

\begin{table*}[!htb]
\setlength\tabcolsep{0pt}
\caption{Performance Comparison of Eye-Tracking Models on 3ET+} \label{tab:perf}
\centering
Comparison of proposed SNN model variants with top-performing ANN models on the 2024 AIS Challenge \cite{wang2024event} and of the 2025 event-based eye-tracking workshop \cite{chen2025event}. ANN models achieve the strongest offline accuracy but rely on computationally heavy components (transformers, state-space blocks, attention) and have been evaluated at 20 Hz. The proposed SNN models operated in a continuous 1 kHz streaming paradigm and retain much of the ANN performance while reducing compute and parameter counts by orders of magnitude. 
$\dag$ denotes self-reported results on the evaluation set.

\smallskip
\begin{tabular*}{\textwidth}{p{-5cm}@{\extracolsep{\fill}}%
lllllcrr}
\toprule
 & Model & {P1} & {P3} & {P5} & {P10} & {Euc. Dist.} & {Params} & {FLOPs} \\
\midrule
 \multirow{11}*{\rotatebox{90}{ANN}} & GTechVision \cite{wang2024event} & 4.16 & 31.70 & 61.08 & 92.26 & 4.94 & 410k & \\
 & CB-ConvLSTM \cite{chen_3et_2023} &  &  & 79.20 & 84.80 & 3.82 & 417k & 1.09T  \\ 
 & PEPNet \cite{ren2024simple} & 7.79 & 49.08 & 80.67 & 97.95 & 3.51 & 640k & 459M \\
 & GoSparse \cite{wang2024event}& 7.32 & 47.97 & 77.20 & 99.00 & 3.51 & 465k & \\
 & MeMo \cite{wang2024event}  & 6.53 & 50.87 & 89.36 & 99.05 & 3.20 & 5.4M & 230M \\
 & CETM \cite{wang2024event} & 23.91 & 83.83 & 96.31 & 99.26 & 2.03 & 7.1M & 2.9G  \\
 & ERVT \cite{wang2024event}  & 28.80 & 87.26 & 94.94 & 98.21 & 1.98 & 150k & 157M\\
 & MambaPupil \cite{wang2024mambapupil}& 33.75 & 90.73 & 97.05 & 99.42 & 1.67 & 8.59M & 2.61T \\
 & TDTracker \cite{ren2025exploring}  & & 91.20\dag & 97.20\dag & 99.20\dag & 1.50 & 3.04M & 265M \\
 & BigBrain \cite{pei2024lightweight}  & 45.50 & 94.58 & 97.79 & 99.00 & 1.44 & 809k & 110.4M \\
 & BRAT \cite{wu2025brat}  & & & \textbf{97.80}\dag & \textbf{99.59}\dag & \textbf{1.14} & 7.1M & 5.8G \\
  \midrule
 \multirow{3}*{\rotatebox{90}{SNN}} & Our (N=512) & 9.59 & 54.76 & 79.10 &  96.25 & 3.66 &   352k & 3.4M \\
& Our (N=256) & 7.27 & 43.51 & 74.85 & 95.45 & 4.10 &  189k & 3.1M \\
& Our (N=128) & 8.57 & 48.80 & 77.27 & 95.85 & 3.90 &  \textbf{107k} & \textbf{2.9M} \\
\bottomrule
\end{tabular*}
\end{table*}

\subsubsection{Fully Event-based ANNs}

%

%


%
Fully event-based approaches process event streams directly using \glspl{ann} without frame-based sensors. However, \glspl{ann} require adapting sparse event data, typically by accumulating events into frame-like representations (e.g., voxel grids or kernel density estimates), which can introduce latency and sacrifice temporal resolution \cite{jiang2024eye}.

Representative methods include Li et al.'s \cite{10168551} segmentation \gls{cnn} with event-based \gls{roi} and E-Gaze's \cite{10416378} kernel density estimation with elliptical fitting. Other approaches such as Yang et al. \cite{yang2023high} use events for frame interpolation before segmentation, further increasing computational burden.

Recurrent architectures have been explored to better model temporal dynamics, as the historical trajectory of the pupil provides valuable information for predicting its future location. Chen et al. proposed a sparse ConvLSTM that reduces operations by 4.7$\times$ compared to conventional \glspl{cnn} \cite{chen_3et_2023}, while Ryan et al. integrated \glspl{gru} into YoloV3 for event-based tracking \cite{ryan2021real}. Recent innovations from the AIS 2024 Challenge \cite{wang2024event} and its subsequent 2025 workshop \cite{chen2025event} include state-space models like MambaPupil for selective temporal modeling \cite{wang2024mambapupil} or the bidirectional transformer BRAT \cite{wu2025brat}, see \Cref{tab:perf}. The state-of-the-art FACET framework demonstrates exceptional eftablficiency with 0.2 pixel error by directly regressing pupil parameters \cite{ding2025facet}.


%

%
Despite these advances, most methods still rely on intermediate frame representations, limiting their temporal resolution and efficiency. Additional limitations include sensitivity to noise, special auxiliary hardware  and dependency on synthetic training data, restricting real-world applicability.

\subsubsection{Fully Event-based Networks}


Fully neuromorphic approaches represent the most integrated solution, processing sparse event streams directly with \glspl{snn} to create a synergistic, event-driven pipeline. This paradigm eliminates the conversion to frames, leveraging the temporal dynamics of \glspl{snn} for efficient feature extraction and enabling lower latency and power consumption.

Key implementations demonstrate different strategies. Retina \cite{bonazzi2024retina} achieved a 3.24px tracking error with <5 mW and $\sim$6 ms latency on neuromorphic hardware. Jiang et al. \cite{jiang2024eye} developed an \gls{snn} capable of self-adjusting its temporal resolution down to 0.081 ms, offering high adaptability but with a computationally intensive architecture. In contrast, Stoffregen et al. \cite{stoffregen2022event} pursued a model-based strategy, tracking corneal glints at 1 kHz, although this method is dependent on controlled illumination.
Despite their promise, these systems reveal persistent trade-offs. \gls{snn} models face challenges with continuous tracking due to neuron state resets on hardware, while other approaches suffer from computational overhead or environmental dependencies. These limitations highlight the need for a robust and efficient neuromorphic solution, a gap the current work aims to address.

\section{Methods}

\subsection{Data}

The experiments in this work use the 3ET+ dataset, a benchmark for event-based eye tracking published as part of the AIS 2024 Event-based Eye Tracking Challenge \cite{wang2024event} and its subsequent 2025 iteration \cite{chen2025event}.
This dataset comprises real recordings from 13 participants, recorded in 2 to 6 sessions and captured using a DVXplorer Mini event camera \cite{dvxplorer-mini}. During the recordings, subjects performed five distinct classes of eye activities: random movements, saccades, reading text, smooth pursuit, and blinks.

The dataset provides a real raw event stream. Each event is represented as a tuple
$(x_i, y_i, t_i, p_i)$, where $x_i$ and $y_i$, are the spatial coordinates with a resolution of 640$\times$480 pixels, $t_i$ is the timestamp with a temporal resolution of 200 $\mu s$, and $p_i$ is the polarity indicating the sign of the brightness change. 
The ground-truth annotations are provided at a frequency of 100 Hz. For each timestamp, the label consists of: 1) the spatial coordinates $(x, y)$ of the pupil center, and 2) a binary value indicating whether the eye was blinking.

For the AIS 2024 Challenge \cite{wang2024event}, the primary evaluation metric on the Kaggle leaderboard was $P$-accuracy (specifically P10-accuracy). A prediction was considered correct if the Euclidean distance to the ground truth was within 10 pixels. Additional metrics, including P-accuracy for stricter tolerances ($P=\{1,3,5\}$) and the $L2$ error, were also used for analysis. Notably, the 2025 iteration \cite{chen2025event} updated its primary metric to mean pixel error (mean Euclidean distance) to better differentiate between high-performing models.

\begin{table}[!htb]
    \caption{Proposed Network}
    \label{tab:net}
    \justifying
     The dimension of depthwise (DW) and pointwise (PW) convolution was chosen to match the dimensions used by CETM and MambaPupil. Convolutional layers have no bias and were padded to retain input dimensions. Hidden layers have fixed decay values that are initialized uniformly between 0.9 and 1. The output is the membrane potential of the final LIF layer with a learnable $\beta$, initialized to 0.9. Output channel $N$ is a hyperparameter, with $N \in (128,\ 256,\ 512)$. \\ \\ 
    \centering
    \begin{tabular}{rlcc}
    \toprule
         &Name & Dimensions & \#Params \\
         \midrule
         \multirow{5}{*}{\hspace{-1em}$\left.\begin{array}{l}
                \\
               \rotatebox{90}{Conv1} \\
                \\
                \end{array}\right\lbrace$}  & DW Conv & $2 \times 2 \times 7 \times 7$ & 98 \\
         &IN, ReLU & $2$ &   \\
         &PW Conv & $2 \times 32 \times 1 \times 1$ & 64 \\
         &IN, ReLU & $32$ &   \\
         &Pool & $3 \times 3$ &  \vspace{.2em}  \\
         \multirow{5}{*}{\hspace{-1em}$\left.\begin{array}{l}
                \\
               \rotatebox{90}{Conv2} \\
                \\
                \end{array}\right\lbrace$} & DW Conv & $32 \times 32 \times 5 \times 5$ & 800 \\
         &IN, ReLU & $32$ &   \\
         &PW Conv & $32 \times 128 \times 1 \times 1$ & 4096  \\
         &IN, ReLU & $128$ &   \\
         &Pool & $3 \times 3$ & \vspace{.2em}   \\
         \multirow{5}{*}{\hspace{-1em}$\left.\begin{array}{l}
                \\
               \rotatebox{90}{Conv3} \\
                \\
                \end{array}\right\lbrace$} & DW Conv & $128 \times 128 \times 5 \times 5$ & 3200  \\
         &IN, ReLU & $128$ &   \\
         &PW Conv & $128 \times N \times 1 \times 1$ & 128N  \\
         &IN, ReLU & $N$ &   \\
         &Pool & $4\times4$ &  \vspace{.2em} \\
         &Flatten \\
         &LIF1 &  $2N\times256$ &  512N + 256 \\
         &LIF2 &  $256\times64$ & 16,448  \\
         &LIF3 &  $64\times2$ & 132 \\
         \midrule
         &\textbf{Total} &   & 640N + 25,094 \\
         \bottomrule
    \end{tabular}
\end{table}

\newpage

\subsection{Implementation Details}

All neuromorphic models were implemented using PyTorch \cite{DBLP:journals/corr/abs-1912-01703} and snnTorch \cite{eshraghian2023training}, with experiments conducted on a single NVIDIA L40S GPU. Raw events were discretized into 1 ms bins, matching the target 1 kHz inference frequency. Within each bin, polarities were aggregated and spatially downsampled to 80$\times$60 pixels to reduce computational load and match the effective resolution used in prior work. 

Ground truth labels were upsampled from 100 Hz to 1 kHz using cubic B-spline interpolation to align with the event windows. This interpolation smooths the trajectory without introducing visible artifacts; visual inspection confirms sub-pixel deviations and avoids discontinuities between interpolation regions. Given the emphasis on low-power decoding rather than fine positional precision, this approximation is deemed acceptable with the benefit that it allows comparison to other reference solutions on the dataset.

To improve model robustness and prevent overfitting, the data was augmented during training. Spatial augmentations included horizontal and vertical flips and shifts. Temporal augmentations included temporal shifts and flips; in the latter case, polarities were inverted to maintain physical consistency. Additionally, we employed Event-Cutout \cite{wang2024mambapupil}, randomly masking spatio-temporal regions to encourage robustness.

Training used the Adam optimizer with a learning rate of \num{1e-3}. The loss function combined position and velocity terms, $L=L_{pos}+L_{vel}$,
where $L_{pos}$ is the framewise MSE of pupil positions and 
$L_{vel}$ is the MSE of first-order temporal differences. The velocity term encourages temporal smoothness and reduces jitter in high-frequency predictions. 

Following the AIS 2024 protocol \cite{wang2024event}, blink periods were excluded from accuracy reporting. Model performance is reported using the standard train-validation-test paradigm, with 31 training, 9 validation and 11 testing sessions. Final results on the test set were reported for the epoch with peak validation performance.
Models were trained on 450 ms sequences (450 timesteps with 1 ms bins) with a 1 ms stride across 40 epochs (\Cref{fig:pipeline}). While \glspl{snn} operate online at inference time, shorter windows proved detrimental during training. Due to the increased sparsity at 1 ms, many shorter sequences contain few events, forcing the network to infer gaze positions from empty frames. The 450 ms window increases the likelihood of multiple events bursts per sequence, improving convergence. This choice is consistent with top AIS performers \cite{wang2024mambapupil, pei2024lightweight} and is validated by our ablations in \Cref{tab:abl}, which show that networks trained with shorter windows fail to learn.

For real-world assessment, models were evaluated in a continuous streaming mode: validation and test sessions were fed as uninterrupted streams of 1 ms windows without segmentation or overlap. During inference, the model uses no temporal context beyond the membrane potentials maintained by the \gls{lif} neurons, making it suitable for 1 kHz real-time operation. 

\begin{table*}[!htb]
\caption{Efficiency comparison with Retina} \label{tab:eff}
\justifying
Efficiency comparison between our neuromorphic models and Retina. FLOPs include convolutional operations and \gls{lif} membrane updates. Sparse FLOPs reflect expected runtime cost given observed firing rates. Latency is projected according to the nature of neuromorphic execution, where each layer processes spikes in successive timesteps. 
$^1$ Sparse FLOPs use the dynamic firing rate from \cite{bonazzi2024retina}. $^2$ End-to-end on Synsense Speck \cite{yao2024spike}. $^3$ Power is projected using the reported energy consumption for the effective arithmetic and memory operations of the neuromorphic chip SENeCA \cite{tang2023open}.

\smallskip
\begin{tabular*}{\textwidth}{@{\extracolsep{\fill}}%
lrrccccc}
\toprule
Variant  & {Params} & {FLOPs} & {Sparse FLOPs}  & {Power} & {Frequency} & {Latency}  \\
\midrule
Retina \cite{bonazzi2024retina} & 63k & 6.06M & $\approx$ 2.66M$^1$  & <5 mW$^2$ & $\leq$5 kHz & >5 ms$^2$\\ 
\midrule
Our (N=512) & 353k & 3.4M & 2.5M & 4.9 mW$^3$ & \multirow{3}*{1 kHz} & \multirow{3}*{3 ms} \\
Our (N=256) &  189k & 3.1M & 2.2M & 4.2 mW$^3$ &   & \\
Our (N=128) & 107k  & 2.9M & 2M & 3.9 mW$^3$ &   & \\
\bottomrule
\end{tabular*}
\end{table*}

\subsection{Models}

The proposed architecture adapts the convolutional backbone shared by MambaPupil \cite{wang2024mambapupil}, CETM, and BRAT \cite{wu2025brat}, replacing their recurrent and attention modules with \gls{lif} layers. To reduce parameter count and computational load, all convolutional blocks are replaced with depth-wise separable convolutions (DSC), following the MobileNet design \cite{sifre2014rigid, howard2017mobilenets}.

The proposed architecture comprises three convolutional blocks for spatial feature extraction. Each block follows the pattern: depth-wise (DW) convolution, instance normalization (IN), ReLU activation, pointwise (PW) convolution IN, ReLU, and average pooling. Instance normalization is used over Batch Normalization, due to its stability in streaming inference, where batch statistics are unavailable. The dimensions of each depth-wise separable block and the subsequent \gls{lif} layers are summarized in \Cref{tab:net}. 

The flattened convolutional features are passed into a stack of 3 dense \gls{lif} layers, replacing the original temporal modules. Hidden \gls{lif} layers use fixed decaying values sampled uniformly from $\beta \sim U(0.9,1)$, providing a range of temporal decay constants and enabling multiple timescales of integration. The output \gls{lif} layer uses a learnable decay parameter, initialized to $\beta_{out} = 0.9$. 
The membrane potentials of the output \gls{lif} neurons, rather than their spike outputs, are used for regression. This produces continuous predictions while maintaining spike-based computation in hidden layers. Membrane potentials are initialized uniformly and reset by subtraction when thresholds are crossed. 
Although neuronal firing rates are not explicitly constrained, overall sparsity emerges naturally from the high input sparsity and \gls{lif} dynamics.

\section{Experimental Results}


\subsection{Comparison with ANN-Based 3ET+ Models}

\Cref{tab:perf} summarizes performance relative to the top \gls{ann} models from the Event-based Eye Tracking 2024 Challenge and the 2025 Workshop. These \gls{ann} architectures, ranging from state-space models (MambaPupil) to attention-driven designs (BRAT), achieve the strongest offline accuracy on 3ET+, but are typically trained at 100Hz and evaluated at 20Hz, and rely on computational mechanisms, such as, transformers, state-space blocks and attention, and do not scale to milliwatt-level or kHz-rate deployment. 

The proposed spiking variants achieve 3.7-4.2px mean error across test sessions. This is 1-2px gap to the top \gls{ann} models (1.1 - 2.0px). This accuracy drop is unsurprising given that heavy temporal modelling blocks are intentionally removed and replaced with lightweight neuromorphic layers. Despite this, the SNN variants preserve a substantial fraction of \gls{ann} performance while operating under drastically reduced computational budgets. 

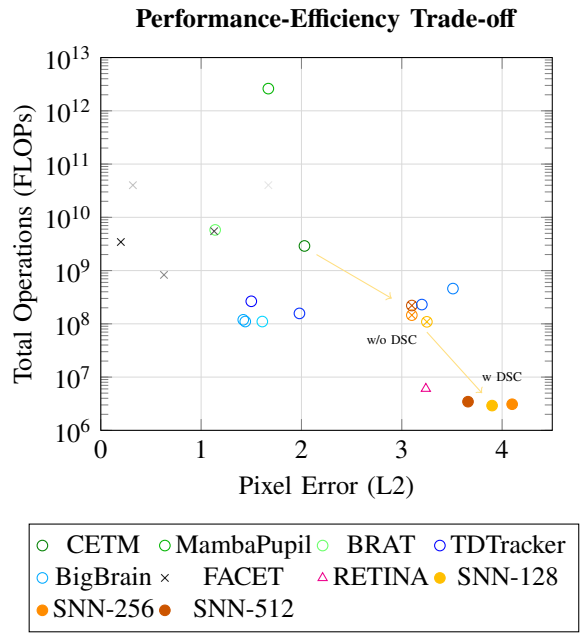
\begin{figure}[!htb]
  \centering

\begin{tikzpicture} 
[spy/.style={%
        draw,blue,
        line width=1pt,
        circle,inner sep=0pt, dotted,
        }]
    
\begin{axis}[
    title={\textbf{Performance-Efficiency Trade-off}},
    width=\linewidth,
    xlabel=Pixel Error (L2), 
    xmin=0, xmax=4.5, 
    ylabel=Total Operations (FLOPs),
    ymode=log,
    ymin=1e6, ymax=1e13, 
    log basis y={10},
    legend style={at={(0.45,-0.25)},anchor=north},
    y label style={at={(axis description cs:-.12,0.9)}, rotate=0, anchor=south east},
    legend columns=4,
    grid, 
    grid style={line width=.1pt, draw=gray!30},
]

\definecolor{amber}{RGB}{255,191,0}
\definecolor{darkorange}{RGB}{255,140,0}
\definecolor{burntorange}{RGB}{204,85,0}

\definecolor{green1}{RGB}{0,128,0}
\definecolor{green2}{RGB}{0,180,0}
\definecolor{green3}{RGB}{102,255,102}

\definecolor{grad1}{RGB}{0,0,255}
\definecolor{grad2}{RGB}{0,42,255}
\definecolor{grad3}{RGB}{0,85,255}
\definecolor{grad4}{RGB}{0,127,255}
\definecolor{grad5}{RGB}{0,170,255}
\definecolor{grad6}{RGB}{0,212,255}
\definecolor{grad7}{RGB}{0,255,255}

\definecolor{grey1}{RGB}{30,30,30}
\definecolor{grey2}{RGB}{85,85,85}
\definecolor{grey3}{RGB}{140,140,140}
\definecolor{grey4}{RGB}{195,195,195}
\definecolor{grey5}{RGB}{230,230,230}

\draw[->,amber!50!white] (axis cs: 2.15, 2000000000) -- (axis cs: 2.9, 300000000);  
\draw[->,amber!50!white] (axis cs: 3.25, 70000000) -- (axis cs: 3.8, 5000000);  


\addplot+[green1, only marks, mark=o] coordinates {
    (2.03, 2900000000)
};
\addlegendentry{CETM};

\addplot+[green2, only marks, mark=o] coordinates {
    (1.67, 2610000000000)
};
\addlegendentry{MambaPupil};

\addplot+[green3, only marks, mark=o] coordinates {
    (1.14, 5800000000)
};
\addlegendentry{BRAT};

\addplot+[grad1, only marks, mark=o] coordinates {
    (1.50, 265000000)
};
\addlegendentry{TDTracker};

\addplot+[grad5, only marks, mark=o] coordinates {
    (1.44, 110400000)
};
\addlegendentry{BigBrain};

\addplot+[grey1, only marks, mark=x] coordinates {
    (0.20, 3440000000)
};
\addlegendentry{FACET};

\addplot+[magenta, only marks, mark=triangle] coordinates {
    (3.24, 6060000)
};
\addlegendentry{RETINA};

\addplot+[amber, only marks, mark=*, mark options={solid, fill=amber}] coordinates {
    (3.9, 2915746)
};
\addlegendentry{SNN-128};

\addplot+[darkorange, only marks, mark=*, mark options={solid, fill=darkorange}] coordinates {
    (4.1, 3093547)
};
\addlegendentry{SNN-256};

\addplot+[burntorange, mark options={solid, fill=burntorange}, mark=*, only marks] coordinates {
    (3.66, 3453739)
};
\addlegendentry{SNN-512};

\addplot+[amber, mark=otimes] coordinates {
    (3.25, 109149891)
};

\addplot+[darkorange, mark=otimes] coordinates {
    (3.10, 145612981)
};

\addplot+[burntorange, mark=otimes] coordinates {
    (3.1, 220623522)
};

\addplot+[grad2, only marks, mark=o] coordinates {
    (1.98, 157000000)
};

\addplot+[grad3, only marks, mark=o] coordinates {
    (3.2, 230024000)
};

\addplot+[grad4, only marks, mark=o] coordinates {
    (3.51, 459000000)
};

\addplot+[grad5, only marks, mark=o] coordinates {
    (1.42, 119075840)
};

\addplot+[grad6, only marks, mark=o] coordinates {
    (1.61, 110390000)
};

\addplot+[grey5, only marks, mark=x] coordinates {
    (1.67, 40190000000)
};

\addplot+[grey4, only marks, mark=x] coordinates {
    (0.32, 40110000000)
};

\addplot+[grey3, only marks, mark=x] coordinates {
    (0.63, 830000000)
};

\addplot+[grey2, only marks, mark=x] coordinates {
    (1.13, 5490000000)
};

\node[] at (axis cs: 2.9,50000000) {\tiny w/o DSC};
\node[] at (axis cs: 4,10000000) {\tiny w DSC};

\end{axis}
\end{tikzpicture}

\caption[Performance-efficiency tradeoff]{Pixel error is shown against total FLOPs on a log scale. Circles denote models trained on 3ET+, crosses on EV-Eye, and triangles on INI-30. Performance is taken from \cite{wang2024event, chen2025event}, \cite{ding2025facet} and \cite{bonazzi2024retina}, respectively. Notably, 3ET+ uses a spatial resolution of 80$\times$60 compared to 64$\times$64 of EV-Eye and INI-30. Our spiking variants (solid circles) retain acceptable accuracies while operating at orders-of-magnitude lower compute. These models reach similar efficency as Retina while offering greater architectural flexibility. Ideal models lie toward the lower-left corner (low error, low compute). 
}
\label{fig:eff}
\end{figure}

The efficiency advantages are shown in \Cref{fig:eff}. We observe a $20\times$ parameter reduction and $850\times$ floating point operations (FLOPs) reduction compared to CETM, the ANN variant most similar to the neuromophic architectures proposed here. FLOP counts comprise convolutional operations as well as dense operations, membrane updates and decays in LIF layers. Even compared against more efficient \gls{ann} models (e.g, BigBrain) the neuromorphic variant reduces FLOPs by at least $30\times$ and $85\%$ of the parameters. This reduction stems from three factors: (i) removal of attention and recurrence, (ii) depthwise-separable convolutions, and (iii) the sparsity and recurrence inherent to \gls{lif} neurons. 

This architectural efficiency enables projected power consumption of 3.9-4.9 mW, calculated using the SENeCA neuromorphic processor energy model \cite{tang2023open} (1.4 pJ per arithmetic operation, 3.7 pJ per data load). The projected 3.9-4.9 mW range achieved by the neuromorphic models puts them on par with Retina (2.89-4.8 mW) \cite{bonazzi2024retina} while maintaining continuous 1kHz inference. This is more than 50$\times$ the temporal resolution of the operating frequency of the original \gls{ann} models. 

Collectively, these results show that a substantial portion of ANN accuracy performance can be preserved under the  constraints of neuromorphic systems, though \glspl{ann} remain superior where power and latency are not limiting factors. For wearable or embedded settings where power and latency are strict constraints, the \gls{snn} variants offer an attractive solution. 

\subsection{Comparison with Neuromorphic Eye-Tracking Systems}

Retina \cite{bonazzi2024retina} represents the state-of-the-art for end-to-end neuromorphic eye tracking, providing a critical baseline for accuracy, power and latency on physical hardware. Retina is specifically engineered for energy efficiency, achieving 3.24px pupil tracking error on the INI-30 dataset, at 
measured power consumption of 2.89-4.8 mW and latency of 5.57-8.01 ms on the Synsense Speck neuromorphic processor \cite{yao2024spike}. 

A direct numerical comparison of accuracy is constrained by differences in evaluation datasets (INI-30 vs. 3ET+) and input resolution ($64\times64$ vs. $80\times60$). However, the SNN variants presented here achieve a comparable performance bracket of 3.7-4.1$\times$ on 3ET+, while hardware projections indicate a similar power envelope of 3.9-4.9 mW. This demonstrates that the presented approach, converting high-performance ANN backbones, can reach the same critical accuracy-efficiency trade-off as a model like Retina, which was designed from the ground up. 

The key distinction lies in architectural philosophy and demonstrated flexibility. Retina employs a highly specialized, shallow convolutional \gls{snn} front-end paired with a fixed, non-spiking temporal weighted-sum filter, a design tightly coupled to the constraints of its target hardware (Speck). In contrast, the work presented here demonstrates a modular conversion pathway. By systematically replacing recurrent and attention blocks in top-performing \glspl{ann} (e.g., MambaPupil, CETM) with LIF layers, 
complex temporal modelling can be efficiently offloaded to an \gls{snn}'s inherent dynamics without hardware-specific architectural specialization. This provides a more generalizable blueprint for migrating other event-based vision architectures to the neuromorphic domain. 

Latency is projected according to the nature of neuromorphic execution, where each consecutive spiking layers is processed subsequently. Given Retina's 8 layers and a measured latency of 5.57-8.01 ms, this projection is conservative. The proposed  model's pipelined architecture, with three successive \gls{lif} layers, yields a theoretical latency of 3 ms, which is lower than Retina's measured latency. It is crucial to note that this is a projection, whereas Retina's figure is a hardware measurement; end-to-end deployment is required for a definitive comparison. Nevertheless, this neuromorphic architecture is inherently suited for continuous, real-time tracking of saccadic movements. 



\section{Discussion}

This work demonstrates that spiking neural networks can achieve the critical balance needed for real-time eye tracking in wearable systems: viable accuracy with extreme efficiency. By converting recurrent \gls{ann} architectures into lightweight, \gls{lif}-based \glspl{snn}, we reduce computational demands by three orders of magnitude, while maintaining acceptable performance on the 3ET+ benchmark. Specifically, we reduce computational costs by a factor of $30$-$1000\times$ and compress model size by 22-44$\times$, all while sustaining an accuracy of 3.7-4.1px. Crucially, the evaluation, conducted in continuous 1 kHz operation with architecture-based power projections, confirms that these models are deployable on neuromorphic hardware for wearable eye tracking systems where milliwatt-level power consumption is advantageous.

The performance comparison reveals fundamentally different design priorities between \gls{ann} and neuromorphic approaches. While \glspl{ann} can achieve superior offline accuracy when evaluated at 20 Hz sampling rate, they often rely on hardware that is too power-hungry or bulky for wearable eye trackers. In contrast, the presented neuromorphic approach focuses on efficient, continuous 1 kHz tracking within stringent power budgets. This enables the capture of fine-grained eye dynamics, such as micro-saccades, with the high temporal and spatial accuracy needed for user authentication or evaluating attention. This trade-off underscores a limitation in conventional benchmarks, that prioritize offline accuracy without accounting for the operational constraints of always-on wearable devices. 

\begin{table}[!hb]
\caption{Ablation study results} \label{tab:abl}
\centering
Impact of architectural and training choices on model performance on the validation set. Notably, there are significant performance differences to the testing set. \\

\smallskip
\begin{tabular*}{\columnwidth}{@{\extracolsep{\fill}}%
lccccc}
\toprule
Variant  & {P5} & {P10} & {Euc. Dist.} & {Params} & {FLOP}  \\
\midrule
 \textit{Benchmark} & 64.20 & 87.49 & 5.26 & \textbf{352k} & \textbf{3.4M} \\\\

\multicolumn{6}{l}{\textbf{Convolution} - \textit{Depth-wise Separable Convolution (DSC)}} \\
w/o DSC-512 & \textbf{75.70} & \textbf{92.01} & \textbf{4.24} &  2M & 280M \\ 
w/o DSC-256 & 74.19 & 91.36 & 4.44 &  1.1M & 203M \\
w/o DSC-128 & 74.36 & 90.77 & 4.44 &  598k & 164M \\
\\
\multicolumn{6}{l}{\textbf{Window} - \textit{450}} \\
$\rightarrow$ 150 & 61.18 & 84.89 & 5.29 & \multicolumn{2}{c}{\textit{unchanged}}\\\\
\multicolumn{6}{l}{\textbf{Stride} - \textit{1}} \\
$\rightarrow$ 10 & 52.27 & 86.20 & 6.07 & \multicolumn{2}{c}{\textit{unchanged}}\\
$\rightarrow$ 50 & 42.04 & 76.52 & 7.65 & \multicolumn{2}{c}{\textit{unchanged}}\\\\
\multicolumn{6}{l}{\textbf{Loss} - \textit{Velocity Loss}} \\
$\rightarrow$ Spatial \\ MSE & 59.74 & 84.75 & 5.72& \multicolumn{2}{c}{\textit{unchanged}}\\
\bottomrule
\end{tabular*}
\end{table}

Beyond raw efficiency, the temporal dynamics inherent to LIF neurons provide distinct advantages for processing event-based data streams. Unlike \glspl{ann}, \glspl{snn} naturally maintain state across timesteps, offering built-in smoothing and noise suppression without requiring explicit architectural mechanisms for temporal modeling. This makes them particularly suited for continuous sensor data where timing precision and energy efficiency are critical. These properties make \glspl{snn} highly suited for real-time eye tracking, as demonstrated by the comparable performance between the presented approach and the specialized Retina neuromorphic benchmark.

Looking forward, closing the accuracy gap with state-of-the-art \glspl{ann} will require addressing of the fundamental benchmark mismatches between offline evaluation and real-time operation. Future work should pursue three key directions; First, end-to-end deployment and optimization on neuromorphic hardware is essential to validate power and latency projections and identify potential bottlenecks in real-world operation. Second, other high-performance architectures such as BigBrain and FACET could be adapted to neuromorphic systems, which could potentially achieve better accuracy while maintaining efficiency. Finally, incorporating temporal filtering mechanisms similar to Retina's weighted-sum approach could enhance prediction stability and smoothness, particularly given the strong performance improvements demonstrated in their work. These combined approaches, hardware validation, architectural expansion, and temporal refinement, represent promising pathways for advancing neuromorphic eye tracking toward both high accuracy and extreme efficiency.
\section{Conclusion}

This work demonstrates that event-based eye-tracking architectures can be effectively redesigned as spiking neural networks, achieving significant efficiency gains while maintaining practical accuracy for real-time wearable use. By replacing the recurrent and attention components of high-performing ANN backbones with lightweight LIF layers, we reduce computational cost by $30$-$1000\times$ and compress model size $22$-$45\times$, while sustaining 3.7-4.1px error on the 3ET+ benchmark. 
Using the SENeCA neuromorphic energy model, our most efficient variants operate at an estimated 3.9-4.9 mW with $\approx$3 ms latency at 1 kHz, making them suitable for continuous, always-on eye tracking systems with a tight power and latency constraints. Although an accuracy gap remains relative to the leading ANN models, the presented results show that neuromorphic redesign can preserve much of their performance at a fraction of the computational footprint. 
Overall, this work provides a concrete step toward deployable, low-power, event-driven gaze estimation. It's hoped that the presented findings hope that our findings, and the accompanying open-source implementations, encourage further exploration of neuromorphic architectures for high-speed, energy-constrained eye-tracking systems. 

\section{Funding}
This research was supported through the NimbleAI project, funded via the Horizon Europe Research and Innovation programme (Grant Agreement 101070679), and UKRI under the UK government’s Horizon Europe funding guarantee (Grant Agreement 10039070); the Horizon Europe AIDA4Edge project (Grant Agreement 101160293); and the EPSRC Edgy Organism project (EP/Y030133/1).

\bibliographystyle{./IEEEtran}
\bibliography{./IEEEabrv,./references}

\end{document}